\documentclass{article}

\usepackage[nonatbib,preprint]{neurips_2020}
\usepackage[numbers]{natbib}

\usepackage[utf8]{inputenc} 
\usepackage[T1]{fontenc}    
\usepackage{hyperref}       
\usepackage{url}            
\usepackage{booktabs}       
\usepackage{amsfonts}       
\usepackage{nicefrac}       
\usepackage{microtype}      
\usepackage{xcolor}
\usepackage{xspace}
\usepackage{tabularx}
\usepackage{amsmath,bm,multicol,multirow}
\usepackage{cleveref}
\usepackage[pdftex]{graphicx}
\usepackage{array, makecell}
\usepackage[scientific-notation=true]{siunitx}
\usepackage{subcaption}
\usepackage{chngpage}

\usepackage{ulem}
\newcommand{\wvpp}{wav2vec 2.0}

\newcommand{\wvppbig}{\textsc{Large}}

\newcommand{\vox}{LibriVox}
\newcommand{\libri}{Librispeech}
\newcommand{\libril}{Libri-light}
\newcommand{\voxsz}{LV-60k}
\newcommand{\librisz}{LS-960}
\newcommand{\libriunsz}{LS-860}

\newcommand{\Inp}{\mathcal{X}}
\newcommand{\Feat}{\mathcal{Z}}
\newcommand{\QFeat}{\mathcal{Q}}
\newcommand{\Context}{\mathcal{C}}

\newcommand{\ze}{\mathbf{z}}
\newcommand{\zq}{\mathbf{q}}
\newcommand{\zqt}{\mathbf{\tilde{q}}}
\newcommand{\cc}{\mathbf{c}}

\title{\LARGE Self-training and Pre-training are Complementary\\for Speech Recognition}

\author{%
Qiantong Xu\thanks{Equal contribution.} \And 
Alexei Baevski$^*$ \And 
Tatiana Likhomanenko \And
Paden Tomasello \And
Alexis Conneau \And
Ronan Collobert \And
Gabriel Synnaeve \And 
Michael Auli \AND
Facebook AI Research
}

\begin{document}

\maketitle

\begin{abstract}
Self-training and unsupervised pre-training have emerged as effective approaches to improve speech recognition systems using unlabeled data. 
However, it is not clear whether they learn similar patterns or if they can be effectively combined.
In this paper, we show that pseudo-labeling and pre-training with wav2vec 2.0 are complementary in a variety of labeled data setups.
Using just 10 minutes of labeled data from Libri-light as well as 53k hours of unlabeled data from LibriVox achieves WERs of 3.0\%/5.2\% on the clean and other test sets of Librispeech -- rivaling the best published systems trained on 960 hours of labeled data only a year ago.
Training on all labeled data of Librispeech achieves WERs of 1.5\%/3.1\%.
\end{abstract}

\section{Introduction}
\label{sec:intro}

Speech recognition models trained on labeled speech data has progressed substantially in the recent past~\cite{park2019specaugment,synnaeve2019end,han2020contextnet,gulati2020conformer}.
A drawback of these models is that they require a lot of labeled data to perform well which is usually only available for English and a few other languages.
Therefore, purely supervised training is impractical for the vast majority of the 7,000 languages spoken around the world~\cite{lewis2016ethnologue} which is why there has been a lot of interest in how to better use unlabeled speech data~\cite{liu2018adversarial,baskar2019semisupervised,hsu2020semisupervised}.

This includes classical self-training~\cite{scudder1965probability,yarowsky1995unsupervised,riloff1996auto} which demonstrated strong results~\cite{parthasarathi2019lessons,kahn2020st,synnaeve2019end,xu2020iterative,park2020improved} by pseudo-labeling unannotated audio data and then retraining the final system with the additional labeled data.
Another line of work is pre-training representations on unlabeled speech followed by fine-tuning on labeled data~\cite{oord2018cpc,schneider2019wav2vec,baevski2019vqwav2vec,chung2019apc,jiang2019improving,kawakami2020learning,rivire2020unsupervised,wang2020unsupervised,baevski2020wav}.

In this paper we combine self-training and unsupervised pre-training which are different approaches to leveraging unlabeled data. 
Both achieved excellent results on competitive benchmarks and the central question we explore is whether the two methods are complementary to each other.
Specifically, we build on the recently introduced wav2vec 2.0 model~\cite{baevski2020wav} and the self-training approach of Kahn et al. (2020;~\cite{kahn2020st}) and Xu et al. (2020;~\cite{xu2020iterative}).
We explore training models on the pseudo-labeled data from scratch or by fine-tuning the pre-trained model.
To better understand how complementary the two methods are, we use the same unlabeled data for both.

Experiments on the full \libri{} corpus as well as the low-resource labeled data setups of \libril{} show that self-training and unsupervised pre-training are indeed complementary, a finding that is inline with recent work in natural language understanding~\cite{du2020selftraining}.
In a very low resource setup with just 10 minutes of labeled data and \vox{} as unlabeled data, the combination of wav2vec 2.0 and self-training achieves a WER of 3.0\%/5.2\% on the clean and other test sets of Librispeech, a relative WER reduction of 25\% and 40\% over recent work in pre-training alone~\cite{baevski2020wav}.
Using just the acoustic model without a language model achieves WER 3.7\%/6.5\% - supporting the hypothesis that self-training distills the language model used for pseudo-labeling into the final model. 
When all 960 hours of labeled training data are used we achieve 1.5\%/3.1\% WER on \libri{}.

\section{Background}
\label{sec:format}

\subsection{Unsupervised Pre-training Model}

We experiment with the recently introduced wav2vec 2.0 model of Baevski et al. (2020;~\cite{baevski2020wav}).
This model contains a convolutional feature encoder $f: \Inp \mapsto \Feat$ to map raw audio~$\Inp$ to latent speech representations $\ze_1, \dots, \ze_T$ which are input to a Transformer $g: \Feat \mapsto \Context$ to output context representations $\cc_1, \dots, \cc_T$~\cite{devlin2018bert,baevski2019vqwav2vec,baevski2019effectiveness}.
Each $\ze_t$ represents about 25ms of audio strided by 20ms and the Transformer architecture follows BERT~\cite{vaswani2017transformer,devlin2018bert}.
During training, feature encoder representations are discretized to $\zq_1, \dots, \zq_T$ with a quantization module $\Feat \mapsto \QFeat$ to represent the targets in the objective.
The quantization module uses a Gumbel softmax to choose entries from $G=2$ codebooks with $V=320$ entries each and the chosen entries are concatenated to obtain $\zq$~\cite{jegou2011ieee,jang2016gumbel,baevski2019vqwav2vec}.

The model is trained by solving a contrastive task over masked feature encoder outputs.
At training time, spans of ten time steps with random starting indices are masked.
The objective requires identifying the true quantized latent $\zq_t$ for a masked time-step within a set of $K=100$ distractors $\mathbf{Q}_t$ sampled from other masked time steps: 
$-\log \frac{\exp(sim(\cc_t, \zq_t))}{\sum_{\zqt \sim \mathbf{Q}_t} \exp(sim(\cc_t, \zqt))}$
where $\cc_t$ is the output of the Transformer, and $sim(\mathbf{a}, \mathbf{b})$ denotes cosine similarity.
The objective is augmented by a codebook diversity penalty to encourage the model to use all codebook entries~\cite{dieleman2018challenge}.

\subsection{Self-training Approach}
\label{ssec:subhead}

We adopt the pseudo-labeling strategy of Kahn et al. (2020;~\cite{kahn2020st})  and Synnaeve et al. (2020;~\cite{synnaeve2019end}). 
This first trains an initial acoustic model on the available labeled data and then labels the unlabeled data with the initial model as well as a language model in a step we call pseudo-labeling.
Finally, a new acoustic model is trained on the pseudo-labeled data as well as the original labeled data.

Previous work considered multiple rounds of pseudo-labeling where the labeling step is repeated with each new model to train another model~\cite{xu2020iterative}.
While iterative pseudo-labeling is more accurate, we opt for a single iteration which is computationally less demanding while still enabling us to reason about whether unsupervised pre-training and pseudo-labeling are complementary. 
Another line of work investigated filtering the resulting pseudo-labeled data to match the distribution of the original labeled data~\cite{park2020improved}.
Both methods may improve results and we leave them to future work.

\subsection{Combining the two Approaches}

To combine the approaches, we replace the initial model for pseudo-labeling with a pre-trained model.
The resulting training pipeline is as follows:
we first pre-train a wav2vec 2.0 model on the unlabeled data, fine-tune it on the available labeled data, use the model to label the unlabeled data, and finally use the pseudo-labeled data to train the final model.
In our experiments, we also consider a variant where we fine-tune the original wav2vec 2.0 model on the pseudo-labeled data.

\section{Experimental Setup}
\label{sec:foot}

\subsection{Datasets}
 \label{sec:setup_data}

As unlabeled data for pre-training and self-training we consider the speech audio of the \libri{} corpus (\librisz{};~\cite{panayotov2015librispeech}) without transcriptions containing 960h of audio as well as the audio data of \vox{} (\voxsz{}). 
For the latter we follow the pre-processing of Kahn et al. (2020;~\cite{kahn2020librilight}) resulting in 53.2k hours of audio.
We consider five labeled data setups: 
all 960h of transcribed \libri{}, the train-clean-100 subset comprising 100h, as well as the \libril{} limited resource training subsets of train-10h (10h), train-1h (1h), and train-10min (10min).
We evaluate on the standard \libri{} dev-other/clean and test-clean/other sets.

\subsection{Pre-trained models}

Pre-trained models are implemented in fairseq~\cite{ott2019fairseq} and we obtain them from the public fairseq github repository.\footnote{\url{https://github.com/pytorch/fairseq/tree/master/examples/wav2vec}}
The repository provides fine-tuned models for the five labeled data setups we consider (\autoref{sec:setup_data}).
We experiment with the \wvppbig{} configuration comprising 24 transformer blocks with model dimension 1,024, inner dimension 4,096 and 16 attention heads, comprising a total of about 300M parameters.
The feature encoder contains seven blocks and the temporal convolutions in each block have 512 channels with strides (5,2,2,2,2,2,2) and kernel widths (10,3,3,3,3,2,2), resulting in a receptive field of about 25ms and a stride of about 20ms.
After pre-training on the unlabeled data, this model is fine-tuned on the labeled data using Connectionist Temporal Classification (CTC;~\cite{graves2006connectionist}) and a letter-based output vocabulary.

\subsection{Self-training}
\label{sec:self-training}

We pseudo-label the audio data of either~\librisz{} or~\voxsz{} using \wvpp{} \wvppbig{} fine-tuned on different labeled data splits.
For labeling, we follow the two-pass rescoring procedure of Synnaeve et al. (2020;~\cite{synnaeve2019end}): 
first, we generate a list of candidate transcriptions by combining \wvpp{} and the standard Librispeech 4-gram language model during beam-search with beam 800.
Next, the n-best list is pruned to the 50 highest scoring entries and then rescored with a Transformer LM trained on the \libri{} language corpus~\cite{baevski2018adaptive,synnaeve2019end}.
The Transformer LM has 20 blocks with model dimension 1,280, inner dimension 6,144 and 16 attention heads. 
The n-gram model obtains perplexity 150.3 on the development set and the Transformer language model 49.2.
We found this to be more efficient than directly integrating the Transformer LM into beam search at little loss in accuracy.
Decoding and rescoring hyper-parameters are tuned on dev-other of \libri{} for each experiment using a random parameter search.
The LM weight and the word insertion penalty \cite{synnaeve2019end} is tuned by randomly sampling values in the range of [0, 5] and [-5, 5] over 128 trials.

\subsection{Final Model}
\label{sec:setup_finalmodel}

We follow Synnaeve et al. (2020;~\cite{synnaeve2019end}) and train a Transformer-based sequence to sequence model with log-Mel filterbank inputs after pseudo-labeling using wav2letter++~\cite{pratap2019w2l}.
The encoder uses a convolutional frontend containing 4 layers of temporal convolutions with kernel width 3, followed by 36 Transformer blocks with model dimension $768$, 4 attention heads and feed-forward network (FFN) dimension $3072$~\cite{vaswani2017transformer,synnaeve2019end}.
The model contains about 300M parameters.

We use a 10k word piece output vocabulary computed from the training transcriptions if the whole \libri{} training set is used as labeled data~\cite{kudo2018sentencepiece}. 
Otherwise, we switch to the 5k WP estimated on the train-clean-100 transcriptions~\cite{xu2020iterative,hsu2020semisupervised}.
Language models are incorporated similar to~\autoref{sec:self-training}. 
We use a 4-gram language model and then rescore with a Transformer LM. 
The beam size used in both decoding and rescoring is 50.

\begin{table}[t]
\centering
\caption{WER on the \libri{} dev and test sets for the \libril{} low-resource labeled data setups of 10 min, 1 hour and 10 hours. 
As unlabeled data we use the audio of \libri{} (\librisz{}) or the larger \vox{} (\voxsz{}).
ST (s2s scratch) trains a sequence to sequence model with a word-piece vocabulary on the pseudo-labeled data from random initialization while as ST (ctc ft) fine-tunes \wvpp{} with the pseudo-labels using CTC and a letter-based vocabulary.
All results are with language models at inference time.
}
\label{tab:lowres}
\begin{tabular}{lcrrrr}
\toprule
\multirow{2}{*}{Model} & Unlbld & \multicolumn{2}{c}{dev} & \multicolumn{2}{c}{test} \\
\cline{3-4}\cline{5-6} 
{} & data & clean & other & clean & other \\
\midrule
\midrule
\multicolumn{6}{l}{\textbf{10 min labeled}} \\
Discr. BERT~\cite{baevski2019effectiveness} & \librisz{} & 15.7 & 24.1 & 16.3 & 25.2 \\
\wvpp{}~\cite{baevski2020wav} & \librisz{} & 6.6 & 10.6 & 6.8 & 10.8 \\
\hspace{0.04in}+ ST (s2s scratch) & \librisz{} & 4.1	& 7.0	& 5.0	& 8.1  \\
\hspace{0.04in}+ ST (ctc ft) & \librisz{} & 3.6 & 6.6 & 4.0 & 7.2 \\
\midrule 
\wvpp{}~\cite{baevski2020wav}& \voxsz{} & 5.0 & 8.4 & 5.2 & 8.6 \\
\hspace{0.04in}+ ST (s2s scratch) & \voxsz{} & 2.6 & 4.7 & 3.1 & 5.4  \\
\hspace{0.04in}+ ST (ctc ft) & \voxsz{} & 2.8 & 4.6 & 3.0 & 5.2 \\
\midrule
\midrule
\multicolumn{6}{l}{\textbf{1h labeled}} \\
Discr. BERT~\cite{baevski2019effectiveness} & \librisz{} & 8.5 & 16.4 & 9.0 & 17.6 \\
\wvpp{}~\cite{baevski2020wav} & \librisz{} &  3.8 & 7.1 & 3.9 & 7.6 \\
\hspace{0.04in}+ ST (s2s scratch)  & \librisz{} & 2.9	& 5.6	& 3.4	& 6.6 \\
\hspace{0.04in}+ ST (ctc ft) & \librisz{} & 2.8 & 5.5 & 3.1 & 6.3 \\
\midrule
\midrule
\multicolumn{6}{l}{\textbf{10h labeled}} \\
Discr. BERT~\cite{baevski2019effectiveness} & \librisz{} & 5.3 & 13.2 & 5.9 & 14.1 \\
IPL~\cite{xu2020iterative} & \librisz{} & 23.5 & 25.5 & 24.4 & 26.0 \\
\wvpp{}~\cite{baevski2020wav} & \librisz{} & 2.9 & 5.7 & 3.2 & 6.1 \\
\hspace{0.04in}+ ST (s2s scratch)  & \librisz{} & 2.5	& 5.1	& 3.5	& 5.9 \\
\hspace{0.04in}+ ST (ctc ft) & \librisz{} & 2.6 & 5.2 & 2.9 & 5.7 \\
\bottomrule
\end{tabular}
\end{table}

\begin{table}[t]
\centering 
\caption{
WER on \libri{} when using the clean 100h subset as labeled data or all 960h of \libri{} (cf.~\autoref{tab:lowres}).
Prior work used 860 unlabeled hours (LS-860) but the total with labeled data is 960h and comparable to our setup.
}
\label{tab:highres}
\begin{tabular}{lcrrrr}
\toprule
\multirow{2}{*}{Model} & Unlbld & \multicolumn{2}{c}{dev} & \multicolumn{2}{c}{test} \\
\cline{3-4}\cline{5-6} 
{} & data & clean & other & clean & other \\
\midrule
\midrule
\multicolumn{6}{l}{\textbf{100h labeled}} \\
Discr. BERT~\cite{baevski2019effectiveness} & \librisz{} & 4.0 & 10.9 & 4.5 & 12.1 \\
ST~\cite{kahn2020st} & \libriunsz{} & 5.4 & 19.0 & 5.8 & 20.1 \\
IPL~\cite{xu2020iterative} & \libriunsz{} & 5.0 & 8.0 & 5.6 & 9.0 \\
Noisy student~\cite{park2020improved} & \libriunsz{} & 3.9 & 8.8 & 4.2 & 8.6 \\
\wvpp{}~\cite{baevski2020wav}  & \librisz{} & 2.1 & 4.8 & 2.3 & 5.0 \\
\hspace{0.04in}+ ST (s2s scratch)  & \librisz{} &2.3	&4.6	&2.7	&5.4 \\
\hspace{0.04in}+ ST (ctc ft) & \librisz{} & 2.2 & 4.6 & 2.4 & 5.0 \\
\midrule
IPL~\cite{xu2020iterative} & \voxsz{} & 3.19 & 6.14 & 3.72 & 7.11\\
\wvpp{}~\cite{baevski2020wav} & \voxsz{} & 1.9 & 4.0 & 2.0 & 4.0 \\
\hspace{0.04in}+ ST (s2s scratch) & \voxsz{} & 1.4 & 2.8 & 1.9	& 3.8 \\
\hspace{0.04in}+ ST (ctc ft) & \voxsz{} & 1.7 & 3.2 & 1.9 & 3.6 \\
\midrule
\midrule
\multicolumn{6}{l}{\textbf{960h labeled}} \\
 \multicolumn{5}{l}{\emph{Supervised}} \\
SpecAugment~\cite{park2019specaugment} & - & - & - & 2.5 & 5.8 \\
ContextNet~\cite{han2020contextnet} & - & 1.9 & 3.9 & 1.9 & 4.1 \\
Conformer~\cite{gulati2020conformer} & - & 2.1 & 4.3 & 1.9 & 3.9 \\
\midrule
\multicolumn{5}{l}{\emph{Semi-supervised}} \\
IPL~\cite{xu2020iterative} & \voxsz{} &  1.85 &  3.26 &  2.10 &  4.01 \\
Noisy Student~\cite{park2020improved} & \voxsz{} & 1.6 &  3.4 &  1.7 &  3.4 \\
wav2vec 2.0~\cite{baevski2020wav} & \voxsz{} & 1.6 & 3.0 & 1.8 & 3.3 \\
\hspace{0.04in}+ ST (s2s scratch) & \voxsz{} & 1.1	& 2.7	& 1.5	& 3.1 \\
\hspace{0.04in}+ ST (ctc fine-tune) & \voxsz{} & 1.6 & 2.9 & 1.8 & 3.3 \\
\bottomrule
\end{tabular}
\end{table}

\section{Results}
\label{sec:results}

\subsection{Low-Resource Labeled Data}
\label{sec:libri}

Pre-training has been shown to be very effective in both high- and low-resource labeled training data setups whereas self-training has been most effective when at least a moderate amount of labeled data is available ($\ge$ 100h;~\cite{xu2020iterative,park2020improved}).
To get a sense of whether the combination of both methods can be even more effective, we start with experiments on the \libril{} setups with 10min, 1h and 10h of labeled data. 
For pre-training and pseudo-labeling we use either the 960h of \libri{} without transcriptions or the 53.2k hours of \vox{} (\autoref{sec:setup_data}).
As baseline we consider \wvpp{} pre-trained on \libri{} and fine-tuned on one of the labeled data splits.

We use the publicly available \wvpp{} models to pseudo-label (ST) the unlabeled data and
then evaluate two options to train the final model on the resulting labels:
one is to train a new sequence to sequence model from random initialization with a word-piece vocabulary (s2s scratch) following Synnaeve et al. (2019; \cite{synnaeve2019end}; \autoref{sec:setup_finalmodel}). 
Another option is to fine-tune \wvpp{} on the pseudo-labeled data with CTC and a letter-based vocabulary (ctc ft).

\autoref{tab:lowres} shows that the combination of pre-training and self-training (\wvpp{} + ST) outperforms pre-training alone (\wvpp{}) across all low-resource setups.
It also achieves a very large improvement over iterative pseudo-labeling~\cite{xu2020iterative} in the 10h labeled setup. 
This is because the initial model is much stronger due to pre-training and it is very difficult to train a good supervised-only model on just 10h of labeled data.

With just 10 minutes of labeled data, the combination of pre-training and pseudo-labeling with \vox{} achieves WER 5.2\% on test-other.
Using \libri{} (\librisz{}) as unlabeled data and 10 minutes of labeled data, \wvpp{} + ST  achieves 4.0\%/7.2\% WER on test-clean/other compared to 4.2\%/8.6\% for the best known pseudo-labeling approach~\cite{park2020improved} which uses 100 hours of labeled data.
More unlabeled data leads to large improvements, reducing WER from 4.0\%/7.2\% for \librisz{} to 3.0\%/5.2\% for \voxsz{}, a relative WER reduction of 25-28\%.
But increasing the amount of labeled data without more unlabeled data leads to diminishing returns - an issue we return to in~\autoref{sec:analysis}.
Fine-tuning (ctc ft) generally outperforms from scratch training of a sequence to sequence model with a WP vocabulary (s2s scratch).
This is likely because the model can leverage the pre-trained representations.

\begin{table}[t]
\centering
\caption{WER on \libri{} with and without a language model (LM) for 10 min, and 960h of labeled data and \vox{} as unlabeled data.
}
\label{tab:lm}
\begin{tabular}{lrrrr}
\toprule
\multirow{2}{*}{Model} & \multicolumn{2}{c}{dev} & \multicolumn{2}{c}{test} \\
\cline{2-3}\cline{4-5} 
{} & clean & other & clean & other \\
\midrule
\midrule
\multicolumn{5}{l}{\textbf{10 min labeled}} \\
\wvpp{}~\cite{baevski2020wav} & 5.0 & 8.4 & 5.2 & 8.6 \\
\hspace{0.04in}- LM & 38.3 & 41.0 & 40.2 & 38.7 \\
\midrule
\wvpp{} + ST (s2s scratch) & 2.6 & 4.7 & 3.1 & 5.4  \\
\hspace{0.04in}- LM & 3.3 &	5.9	& 3.7 & 6.5 \\
\wvpp{} + ST (ctc ft) & 2.8 & 4.6 & 3.0 & 5.2 \\
\hspace{0.04in}- LM & 4.2 & 6.9	& 4.3 & 7.2 \\
\midrule
\midrule
\multicolumn{5}{l}{\textbf{960h labeled}} \\
wav2vec 2.0~\cite{baevski2020wav} & 1.6 & 3.0 & 1.8 & 3.3 \\
\hspace{0.04in}- LM &  2.1 & 4.5 & 2.2 & 4.5 \\
\midrule
\wvpp{} + ST (s2s scratch) & 1.1	& 2.7	& 1.5	& 3.1 \\
\hspace{0.04in}- LM & 1.3 &  3.1 &  1.7 &  3.5  \\
\wvpp{} + ST (ctc ft) & 1.6 & 2.9 & 1.8 & 3.3 \\
\hspace{0.04in}- LM & 1.7 & 3.6 & 1.9 & 3.9 \\
\bottomrule
\end{tabular}
\end{table}

\begin{table}[t]
\centering 
\caption{The main driver of performance is the ratio between labeled and unlabeled data. 
We add 8.6 times as much unlabeled data to each labeled setup. Results are on dev-other with an n-gram model and subsets of \librisz{} as unlabeled data.
}
\label{tab:ratio}
\begin{tabular}{lllrr}
\toprule
& labeled & unlab & dev-other & \% change \\
\midrule
\wvpp{} & 10min & 86 min & 12.9 \\
+ ST  & 10min & 86 min & 12.0 & 7\% \\
\midrule
\wvpp{} & 1h & 8.6h & 8.5 \\
+ ST  & 1h & 8.6h & 7.6 & 11\% \\
\midrule
\wvpp{} & 10h & 86h & 6.9 \\
+ ST  & 10h & 86h & 6.5 & 6\% \\
\midrule
\wvpp{} & 100h & 860h & 5.7 \\
+ ST  & 100h & 860h & 5.3 & 7\% \\
\bottomrule
\end{tabular}
\end{table}

\subsection{High-Resource Labeled Data}

Next, we evaluate performance with more labeled data.
We consider the 100h clean subset of \libri{} as well as all 960h of labeled data in \libri{}.
\autoref{tab:highres} shows that \librisz{} as unlabeled data is not enough to outperform the baseline when 100h of labeled data is available.
However, performance improves when using the much larger \voxsz{}, achieving a 10\% relative WER reduction on test-other over \wvpp{}.

When using the full Librispeech benchmark as labeled data, combining \wvpp{} and pseudo-labeling achieves WER 1.5\%/3.1\%.
This result was achieved with a strong sequence to sequence model trained from scratch. 
While less effective than fine-tuning with CTC on smaller setups, a powerful sequence to sequence model excels in this larger setting since the decoder part of the model, which acts in part like a language model, does not overfit.
The lower performance of fine tuning is likely due to CTC not being as competitive as more elaborate sequence to sequence models when a lot of pseudo-labeled data is available~\cite{synnaeve2019end}.

\subsection{Results without a Language Model at Inference Time}

\autoref{tab:lm} shows that combined training models have very good performance even without a language model.
This is because the language model used during pseudo-labeling was partly distilled into the pseudo-labeled data \cite{synnaeve2019end}.
This effect is particularly striking for the 10 min labeled setup without LM where \wvpp{} + ST (s2s scratch) reduces the WER of the baseline (\wvpp{} - LM) by 83\% relative on test-other.
As more labeled data becomes available, the performance of the acoustic model without a language model improves but there is still a clear effect of self-trained models having distilled the language model.
Generally, the sequence to sequence model in the (s2s scratch) setting is better able to distill the language model used at pseudo-labeling time compared to the CTC model used in fine-tuning.

\section{Analysis}
\label{sec:analysis}

We previously saw that improvements decreased with more labeled data (\autoref{sec:libri}).
To better understand this, we perform an experiment on \libri{} where we consider data setups with a fixed ratio between the unlabeled and labeled data.
\autoref{tab:ratio} shows that relative improvements are a function of the amount of unlabeled data relative to the labeled data, rather than the amount of labeled data alone.
\autoref{tab:lowres} showed much larger improvements for the 10 min labeled split but with a fixed ratio of labeled and unlabeled data, the relative improvement is comparable to the 100h labeled setup (\autoref{tab:highres}).

\section{Conclusion}

Unsupervised pre-training and pseudo-labeling are complementary for speech recognition.
This enables building speech recognition systems with as little as 10 minutes of transcribed speech with word error rates that only a year ago were reserved to the best systems trained on 960 hours of labeled data.

\bibliography{refs}
\bibliographystyle{unsrtnat}

\end{document}